\documentclass[10pt,twocolumn,twoside]{IEEEtran}
\usepackage{cite}
\usepackage{amsmath,amssymb,amsfonts}
\usepackage{algorithmic}
\usepackage{graphicx}
\usepackage{textcomp}

\usepackage{hyperref}
\usepackage{epsfig}
\usepackage{graphicx}
\usepackage{amsmath}
\usepackage{amssymb}
\usepackage{multirow}
\usepackage{dsfont}
\usepackage{subfigure}
\usepackage{amsfonts}
\usepackage{epsfig}
\usepackage{epstopdf}
\usepackage{caption}
\usepackage{wrapfig}
\usepackage{bm}
\usepackage{color}
\usepackage{amsmath}
\usepackage{tikz}
\newcommand{\cbet}{\tikz[baseline=-0.6ex]\draw[black,fill=black] (0,0) circle (.3ex);}
\newcommand{\cwor}{\tikz[baseline=-0.6ex]\draw[black] (0,0) circle (.3ex);}

\newcommand{\ie}{\textit{i}.\textit{e}.}

\usepackage{url}

\def\BibTeX{{\rm B\kern-.05em{\sc i\kern-.025em b}\kern-.08em
    T\kern-.1667em\lower.7ex\hbox{E}\kern-.125emX}}

\begin{document}
\title{Ro-SOS: Metric Expression Network for Robust Salient Object Segmentation}

\author{Delu Zeng$^*$, Yixuan He, Li Liu, Zhihong Chen, Jiabin Huang, Jie Chen and John Paisley
\thanks{Delu Zeng, Yixuan He and Zhihong Chen are with School of Mathematics, South China Uinversity of Technology, China.(emails:
dlzeng@scut.edu.cn, Y.He-48@sms.ed.ac.uk, chenzhihong12341234@gmail.com). While Yixuan He is also with School of Mathematics, University of Edinburgh, Scotland, United Kingdom.}
\thanks{Li Liu is with the College of System Engineering, University of National Defense Technology, China. She is also with the Center for Machine Vision and Signal Analysis, University of Oulu, 90014 Oulu, Finland.
(email: li.liu@oulu.fi).}
\thanks{Jiabin Huang is with School of
Information Science and Engineering, Xiamen University, China (email:
Huangjiabin@xmu.edu.cn).}
\thanks{Jie Chen is with School of Electronic and Computer Engineering, Peking University
Peng Cheng Laboratory.(email:chenj@pcl.ac.cn).}
\thanks{John Paisley is with Department of Electrical Engineering, Columbia University, New York, NY 10027, USA (email: jpaisley@columbia.edu)}
\thanks{Corresponding author: dlzeng@scut.edu.cn}
\thanks{This work was supported in part by grants from National Science Foundation of China (No.61571005, No.61811530271), the China Scholarship Council (CSC NO.201806155037), the Science and Technology Research Program of Guangzhou, China (No.201804010429), the Fundamental Research Funds for the Central Universities, SCUT (No.2018MS57).}}
\maketitle

\begin{abstract}
Although deep CNNs have brought significant improvement to image saliency detection, most CNN based models are sensitive to distortion such as compression and noise. In this paper, we propose an end-to-end generic salient object segmentation model called Metric Expression Network (MEnet) to deal with saliency detection with the tolerance of distortion. Within MEnet, a new topological metric space is constructed, whose implicit metric is determined by the deep network. As a result, we manage to group all the pixels in the observed image semantically within this latent space into two regions: a salient region and a non-salient region. With this architecture, all feature extractions are carried out at the pixel level, enabling fine granularity of output boundaries of the salient objects. What's more, we try to give a general analysis for the noise robustness of the network in the sense of Lipschitz and Jacobian literature. Experiments demonstrate that robust salient maps facilitating object segmentation can be generated by the proposed metric. Tests on several public benchmarks show that MEnet has achieved desirable performance. Furthermore, by direct computation and measuring the robustness, the proposed method outperforms previous CNN-based methods on distorted inputs.
\end{abstract}


\section{Introduction}
\label{sec:introduction}
\textcolor{black}{\IEEEPARstart{I}{mage} saliency detection and segmentation is of significant interest in the fields of computer vision and pattern recognition. It aims to simulate the perceptual mechanism of the human visual system by distinguishing between salient regions and background. \textcolor{black}{It is a fundamental problem in CV and has a wide range of applications, including image} segmentation \cite{li2011saliency}, objection detection and recognition \cite{ren2014region}, image quality assessment \cite{li2013color}, \textcolor{black}{visual tracking \cite{zhang2010visual},}
image editing and manipulating \cite{margolin2013saliency}.}
\textcolor{black}{
}

Recent saliency detection studies can be divided into two categories: \textit{handcrafted features based} and \textit{learning-based} approaches. In previous literature, the majority of saliency detection
methods use handcrafted features. Traditional low-level features for such saliency detection models mainly consist of color, intensity, texture edge and structure \cite{2,5,yang2013saliency}. Though handcrafted features with heuristic priors perform well in simple scenes, they are not robust to more challenging cases, such as when salient regions share similar colors to background.

\textcolor{black}{Deep convolutional neural networks (CNNs) \cite{9} have achieved tremendous success in many computer vision problems including image classification \cite{li2016hyperspectral,hu2015deep}, object detection \cite{Liu2018DeepGeneric} and semantic segmentation \cite{GarciaGarcia2018Suvery}, due to their great ability of learning powerful feature representations from data automatically. Likewise, the accuracy of salient detection has been significantly improved and pushed into a new stage by various deep CNN based approaches \cite{li2016visual,wang2016saliency,li2016deep,kuen2016recurrent}. For instance, Wang et al.\cite{SRM} developed a multistage refinement mechanism to effectively combine high-level semantics with low-level image features to produce high-resolution saliency maps. \cite{luo2017non,DHSNet,Amulet} exploited multi-scale convolutional features for object segmentation. Nevertheless, even with great success, there is still plenty room for further improvement. For example, very little work pays attention to the robustness against distorted scenes, while the performance of neural networks are susceptible to typical distortions such as noise \cite{chen2017image}.}

%
 \begin{figure}[!t]
      \subfigure{\includegraphics[width = .32\columnwidth]{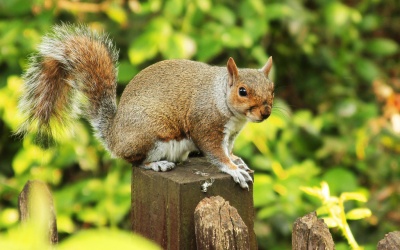}}
      \subfigure{\includegraphics[width = .32\columnwidth]{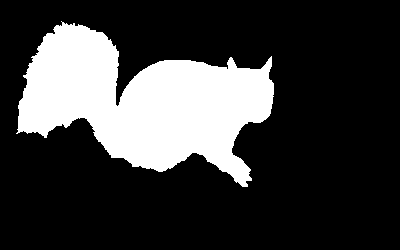}}
      \subfigure{\includegraphics[width =.32\columnwidth]{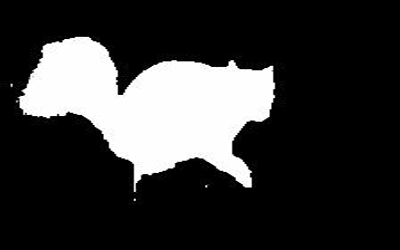}}
      \vspace{-4pt}
      \subfigure[{Image}]{\includegraphics[width = .32\columnwidth]{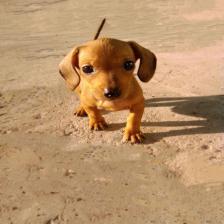}}
      \subfigure[{GT}]{\includegraphics[width = .32\columnwidth]{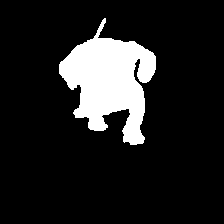}}
      \subfigure[{MEnet}]{\includegraphics[width =.32\columnwidth]{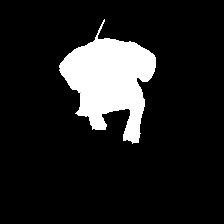}}
      \caption{Saliency segmentation by our algorithm. }\label{fig.result0}
 \end{figure}

\begin{figure*}[t!]
  \includegraphics[width=1\textwidth]{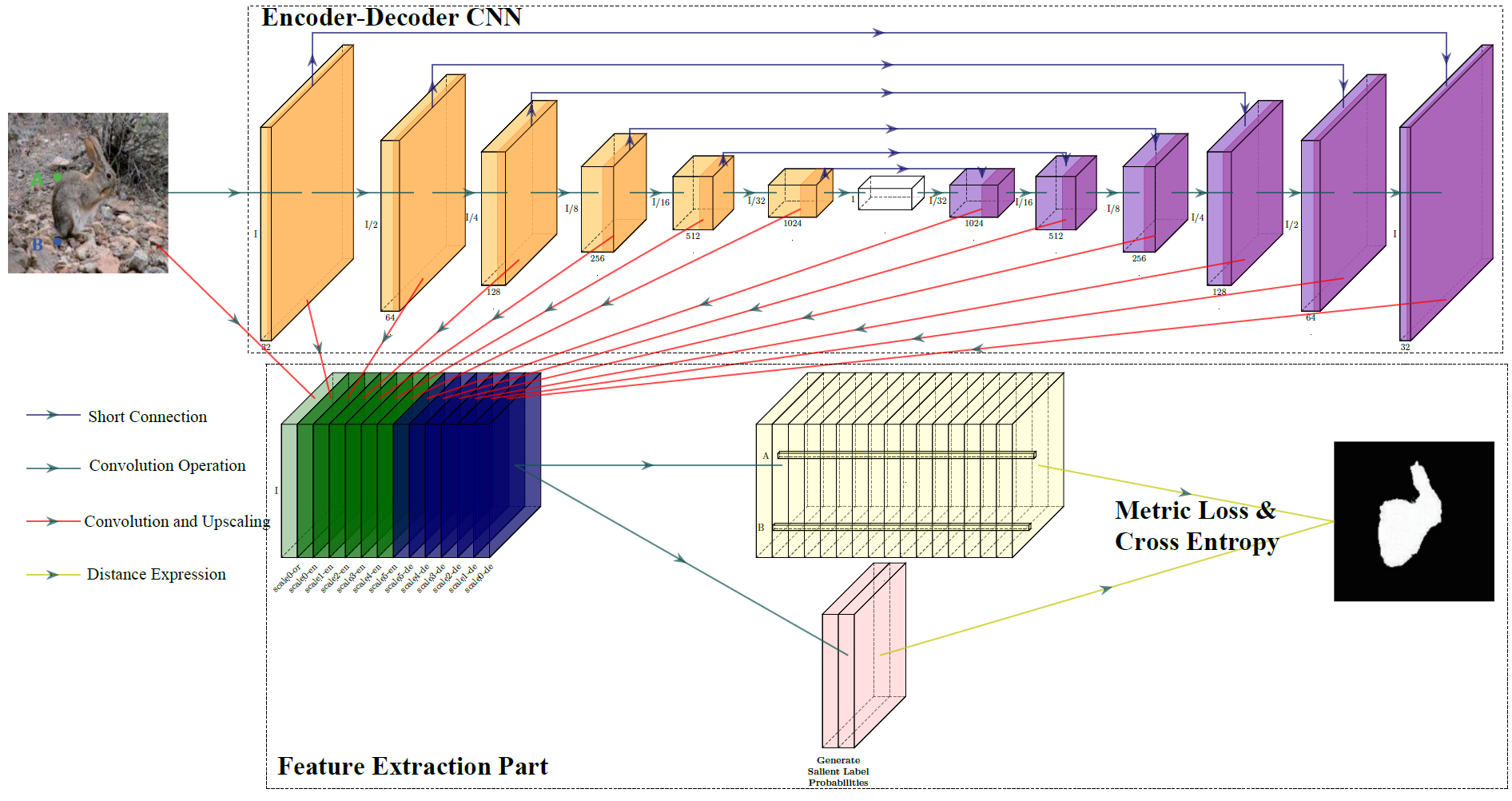}
  \caption{The proposed framework for saliency segmentation.}
  \label{fig.network}
\end{figure*}

\textcolor{black}{ Recently, metric learning has received much attention in computer vision, such as image segmentation \cite{88}, face recognition \cite{hu2014discriminative} and human identification \cite{yi2014deep}, for measuring similarity between objects. Inspired by the metric learning framework, we propose a saliency model that works in a learned metric space.} 
\textcolor{black}{We propose a deep metric learning architecture for saliency segmentation with potentially distorted images.  We use semantic features extracted from a deep CNN to learn a homogeneous metric space. The features are at the pixel level and allow for distinguishing between salient regions and background through a distance measure.  Simultaneously, we introduce a novel metric loss function which is based on metric learning and cross entropy. We also use multi-level information for feature extraction, similar to approaches such as Hypercolumns \cite{hypercolumns} and U-net \cite{U-net}.
We experiment with several benchmark data sets and achieve state-of-the-art level results. Moreover, the proposed model is robust to distorted images.  }

 \textcolor{black}{The rest of this paper is as follows. Section \ref{Related works} introduces related works of saliency segmentation and metric learning. Section \ref{approach} describes details of our model. Finally, Section \ref{experiments} discusses the performance of MEnet model compared to other state-of-the-art saliency segmentation approaches and the robustness to distorted scenes. }

\section{Related works}
\label{Related works}
\subsection{Instance Aware Semantic Segmentation}
Brought about by Fully Convolution Network (FCN)\cite{long2015fully}, CNNs can address semantic segmentation problems and reach state-of-the-art performance, trained in an end-to-end manner, which facilitates following researches in the field. U-Net\cite{U-net} advanced FCN by balancing the number of downsampling and upsampling layers which made the structure symmetric. Evidence shows that semantic segmentation performance benefits from multi-scale information\cite{yu2015multi,chen2016attention,chen2017deeplab,eigen2015predicting,farabet2012learning,pinheiro2014recurrent}. Deeplab\cite{chen2017deeplab} proposed atrous spatial pyramid pooling (ASPP) to leverage the full power of multiple scales.
\subsection{Metric Learning}
In recent years, it has been shown that training a set of data to learn a distance metric can improve performance. Therefore, metric learning is popular in face recognition \cite{guillaumin2009you,hu2014discriminative}, image classification \cite{hastie1996discriminant,zhang2003parametric}, human Re-identification \cite{koestinger2012large}, information retrieval \cite{li2015weakly} and visual tracking \cite{hu2016deep}. Deep learning has achieved much attention, and several metric learning models that are based on deep convolutional neural networks have been proposed. For example, in visual tracking \cite{hu2016deep}, metric learning was used to measure the difference between adjacent frames of the video. In image retrieval \cite{li2015weakly}, using deep metric learning to learn a nonlinear feature space can help to easily measure the relationship of images.

 \textcolor{black}{In recent years, metric learning also has been applied to saliency detection. Lu et al.\cite{li2015adaptive} proposed an adaptive metric learning model based on the global and local information which used the Fisher vector to represent the super-pixel block, then measured the distance of the saliency and background. \cite{you2016salient} employed metric learning to learn the point-to-set metric that can explicitly compute the distances of single points to sets of correlated points, which has the ability to distinguish the salient regions and background.}
\subsection{CNNs-Based Salient Segmentation Approaches}
\textcolor{black}{
Over the past decades, salient segmentation models have been developed and widely used in computer vision tasks. In particular, CNN-based methods have obtained much better performance than all traditional methods which use handcrafted features. The majority of salient segmentation models are based on handcrafted local features, global features or both of them. Most of them prove that using both features performs better than others. For instance, in \cite{li2016deep}, Li et al. proposed deep contrast network that consists of a pixel-level fully convolutional stream and a segment-wise spatial pooling stream, and operates at the pixel level to predict saliency maps. A fully connected CRF is used as refinement. Zhao et al. \cite{MC} combine the global context and local context to train a multi-context deep learning framework for saliency detection. Wang et al. \cite{wang2015deep} used a deep neural network (DNN-L) to learn local patch features and used deep neural network (DNN-G) to obtain the score of salient regions based on the initial local saliency map and global features. Lee et al. \cite{ELD} proposed a unified deep learning framework which utilizes both high level and low level features for saliency detection. In \cite{luo2017non}, Zhang et al. presented a simplified convolutional neural network that combines local and global information and implemented a loss function to penalize errors on the boundary.}

\textcolor{black}{ There are also several works for salient segmentation are based on multi-level features. For example, In \cite{DHSNet}, Liu et al. proposed a deep hierarchical networks for salient segmentation that is first use the GV-CNN to detect salient objects in a global perspective, then HRCNN is considered as refinement method to achieve the details of the saliency map step by step. Zhang et al. \cite{Amulet} presented a multi-level features aggregation network, the deep network first integrates the multi-level feature maps, then it adaptively learns to combine these feature maps and predicted saliency maps. }

\textcolor{black}{In this paper, we proposed a novel a symmetric encoder-decoder CNN architecture for salient segmentation. Our approach differs from the above mentioned methods. Our model ensures that there are enough large receptive fields to obtain much feature information in convolutional operation. Unlike the these methods \cite{Amulet, DHSNet, luo2017non}, we needn't any pre-trained model, and use the different up-sampling method. simultaneously, we construct a effective loss function to predict saliency maps. And in the following section, we will give the details of the proposed model.}

\section{Metric Expression Network (MEnet)}
\label{approach}

We illustrate our model architecture in Figure \ref{fig.network}. An encoder-decoder CNN first generates feature maps at different scales (blocks), which through convolution and up-sampling gives a feature vector for each pixel of an image according to how it maps through the layers. These extracted features are then used in a metric loss and cross entropy function by convolutions for saliency detection as described below.

\subsection{Encoder-decoder CNN for feature extraction}
\textcolor{black}{In SegNet \cite{Segnet} and U-net, the encoder-decoder is used to extraction multi-scale features. We use a similar structure in the proposed model.} Since global information plays an important role in saliency segmentation \cite{wang2015deep}, we use convolutions and pooling layers to increase the receptive field of the model, and compress all feature information into feature maps whose size are $1\times 1$, as shown as the white box in Figure \ref{fig.network}.
Through the decoder module, we up-sample these feature maps and the feature map at each scale represents information at one semantic level. We therefore propose a symmetric encoder-decoder CNN architecture.

\begin{figure}[b]
\centering
\includegraphics[width=1\columnwidth]{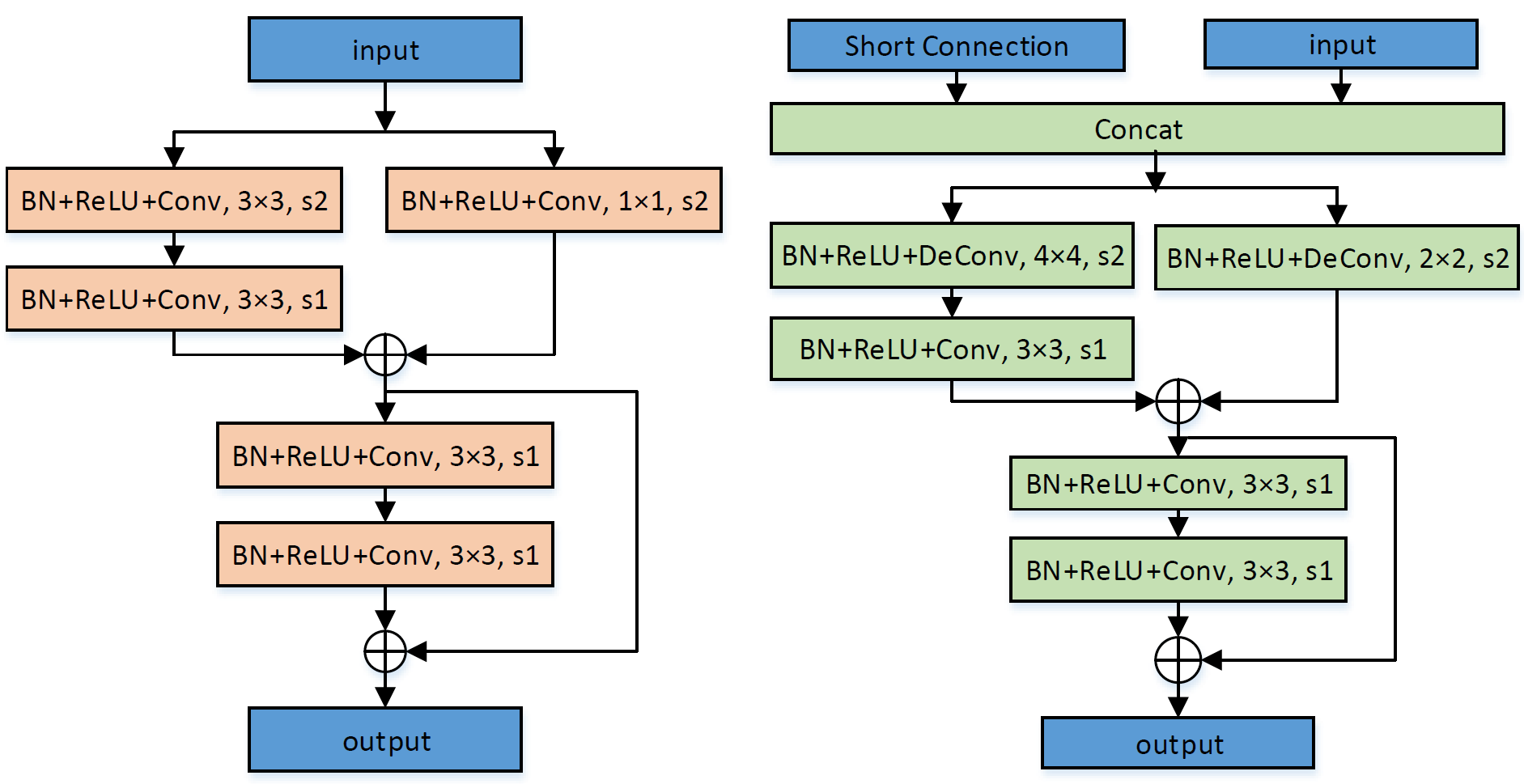}
\caption{Basic encoder (left) and decoder (right) blocks.}
\label{fig.down-up}
\end{figure}

The encoder-decoder network of Figure \ref{fig.network} uses a deep symmetric CNN architecture with short connections as indicated by blue arrows. It consists of an encoder half (left) and a decoder half (right), each block of which has an application of either of the two basic blocks as shown in Figure \ref{fig.down-up}. For encoding, at each down-sampling step we double the number of feature channels using a
convolution with stride 2.
For decoding, each step in the decoder path consists of an up-sampling of the feature map by a deconvolution after being concatenated the input with the short connection, also with
stride 2.
In the decoder path, \textcolor{black}{we concatenate the corresponding feature maps from the encoder path. This part is similar to U-Net. But the difference is that U-net is designed for edge-detection, it  works well even though it crops feature maps (in Fig.1 of U-net) from the encoder path as it doesn't impact edge-detection. While for saliency segmentation, we maintain the size of the feature map to make full use of all the information as its receptive field need to be much larger. We believe that the $1\times1$ size of the feature maps contain rich global information of the original image that can be used by later layers for better prediction.}
Our goal in using a symmetric CNN is to generate different scales of feature maps, which are concatenated to give feature vectors for each corresponding pixel in the
input image that contains multi-scale information across the dimensions. Furthermore, we process some more convolution to balance the dimensional unevenness as described in the following paragraph without doing direct classification.

We ultimately want to distinguish salient objects from background and so want to map image pixels into a feature space where that distance across salient and background regions is large, but within regions is small.
However, previous work in this direction showed that deep CNNs can learn such a feature representation that captures local and global context information for saliency segmentation\cite{MC}. 

Therefore, as it is shown in Figure \ref{fig.network}, we can convert the blocks from the 13 different scales of the encoder-decoder network into a bundle of feature maps as indicated by the {red lines}. That is, in the feature extraction part, each scale generates one output feature map of the same size via a single convolution and up-sampling; while the first ``feature map'' is simply obtained from convolving the original image across its RGB channels.
\textcolor{black}{Though the proposed algorithm may be partially similar to Hypercolumns model, during testing Hypercolumns model takes the outputs of all these layers, upsamples them using bilinear interpolation and sums them for the final prediction.
But the difference is that, within training Hypercolumns model predicts heatmaps from feature maps of different scales by stacking on additional convolutional layers. Hypercolumns is more like DHSNet \cite{DHSNet} which utilizes multi-scale saliency labels for segmentation. In contrast, MEnet upsamples each scale of feature map with the same size during training.}
As the components from these features with
13 scales are uneven to each other, they cannot directly be applied to classification, e.g., assigning any loss function on them. These components from different scales should be balanced and one possible way is to filter these features of 13 dimension by convolution. Consequently in the proposed way, after concatenating the feature maps at each level we further use convolution operation with 16 kernels to generate the final feature map to balance the dimensional characteristics with the
constraint of minimizing the cross entropy (see the following section). In this case, the final feature vector is in $\mathbb{R}^{16}$.
\subsection{Loss function}
\label{sec:loss function}
Most previous work on saliency detection based on deep learning used cross entropy (CE) to optimize the network \cite{li2016deep,luo2017non}. The loss functions is written as follows:
    \begin{equation}
    \begin{aligned}
    {L_{CE}}&(l|\theta_1)=-\frac{1}{N\times |\Omega|}\cdot\\&\sum_{n=1}^{N}\sum_{i=1}^{\Omega}\sum_{y=0}^{1}1\{l{^{(n)}_i}=y\}\ln P(l{^{(n)}_i}=y|\theta_1),
    \end{aligned}
    \end{equation}\label{eq.3}
where $\theta_1$ is the set of learnable parameters of network to influence $l$, $\Omega$ is the pixel domain of the image, $L_{CE}$ denotes the loss for the $n$-th image in the training set, $\bm{1}\{\cdot\}$ is the indicator function; and $y\in \{0,1\}$, where $y=1$ denotes the salient pixel and $y=0$ denotes the non-salient pixel. $P(l{^{(n)}_i}=y|\theta_1)$ is the label probability of the $i$-th pixel predicted by network. In MEnet, we generate $P(l{^{(n)}_i}=y|\theta_1)$ via a convolution with 2 kernels from feature extraction part as shown in Figure \ref{fig.network}.

 Inspired by the metric learning, we also introduce our metric loss function (ML) defined as Equation \ref{eq.5}. In our network, the input is an RGB image whose size is $H\times W\times3$, and all the images are resized to $224 \times 224 \times3$, hence $H=W=I=224$ here. The output is a feature metric space generated by 16-kernels convolution in Figure \ref{fig.network}, and the size is $H\times W\times C$ (in our method we set $C=16$). Each pixel in the $H\times W$ image corresponds to a C-dimension vector in the salient feature maps. The metric loss function is defined as:
    \begin{equation}\label{eq.5}
    \begin{aligned}
    L_{ML}(f|\theta_2) = \frac{1}{N\times |\Omega|}&\sum_{n=1}^{N}\sum_{i=1}^{\Omega} \big(\frac{1}{\left| set^+  \right|}\cdot\\\sum_{k \in set^{+}}\| {f_{i}^{(n)} - f_ k ^{(n)}} \| _2^2 -
    &\frac{1}{{\left| {set^- } \right|}}\sum_{k \in set^{-}}\| {f_{i}^{(n)} - f_ k ^{(n)}} \| _2^2\big)
    \end{aligned}
    \end{equation}

where $\theta_2$ is the set of learnable parameters of network to influence $f$, $f_{i}^{(n)}$ denotes the feature vectors corresponding to the pixel in the $n$-th image of the training set. We denote $k \in set^{+}$ (or $k \in set^{-}$), s.t., $\Omega=set^{+}\cup set^{-}$ , by meaning that $f_k^{(n)}$ is the positive (negative) feature vector of $f_{i}^{(n)}$, respectively. That is, $f_i^{(n)}$ and $f_k^{(n)}$  are from the same region (salient or non-salient), otherwise, $f_k^{(n)}$  is from a different region with respect to $f_{i}^{(n)}$ . We use Euclidean distance to calculate the distance between two feature vectors.

This loss function (\ref{eq.5}) seeks to find an encoder-decoder network that enlarges the distance between any pair of feature vectors from different regions, and reduces the distance from the same region. In this way, the two region is expected to be homogenous by themselves. Then by trivial deduction, it is  equivalent to
\begin{equation}\label{eq.6}
\begin{aligned}
{L_{ML}^{*}}(f|\theta_2) = \frac{1}{N\times |\Omega|}\sum_{n=1}^{N}\sum_{i=1}^{\Omega} \big(\| {f_{i}^{(n)}- \bar{f_{+}} ^{(n)}} \| _2^2\\ - \| {f_{i}^{(n)} - \bar{f_{-}}^{(n)}} \| _2^2\big)
\end{aligned}
\end{equation}
where we average all $f_ k ^{(n)}$ in Equation \ref{eq.6} to get $ \bar{f_{+}} ^{(n)}$ and $\bar{f_{-}}^{(n)}$. That is $ \bar{f_{+}} ^{(n)}$ is the mean of all positive pixels from a single image, while $\bar{f_{-}}^{(n)} $ corresponds to all negative pixels. Intuitively,  Equation \ref{eq.6} enforces that the feature vectors extracted from the same region be close to the center of that region while keeping away from the center of other region in salient feature space. In this case, we can obtain a more robust distance evaluation between the salient object and background. We also add a second cross entropy loss function as a constraint which shares the same network architecture to the objective function and empirically have noticed that the combined results were significantly better than only using either the metric or cross entropy losses. Therefore, our final loss function is defined as below:
\begin{equation}\label{eq.7}
{L_{MEnet}}(f,l|\theta) = {L_{ML}^{*}}(f|\theta_2) + \lambda {L_{CE}}(l|\theta_1 ),
\end{equation}
where $\theta=\theta_1 \cup \theta_2$ and $\lambda$ is set to 1 in our experiment.

\subsection{Semantic distance expression}

\begin{figure}[t]
\centering
\includegraphics[width=0.8\columnwidth]{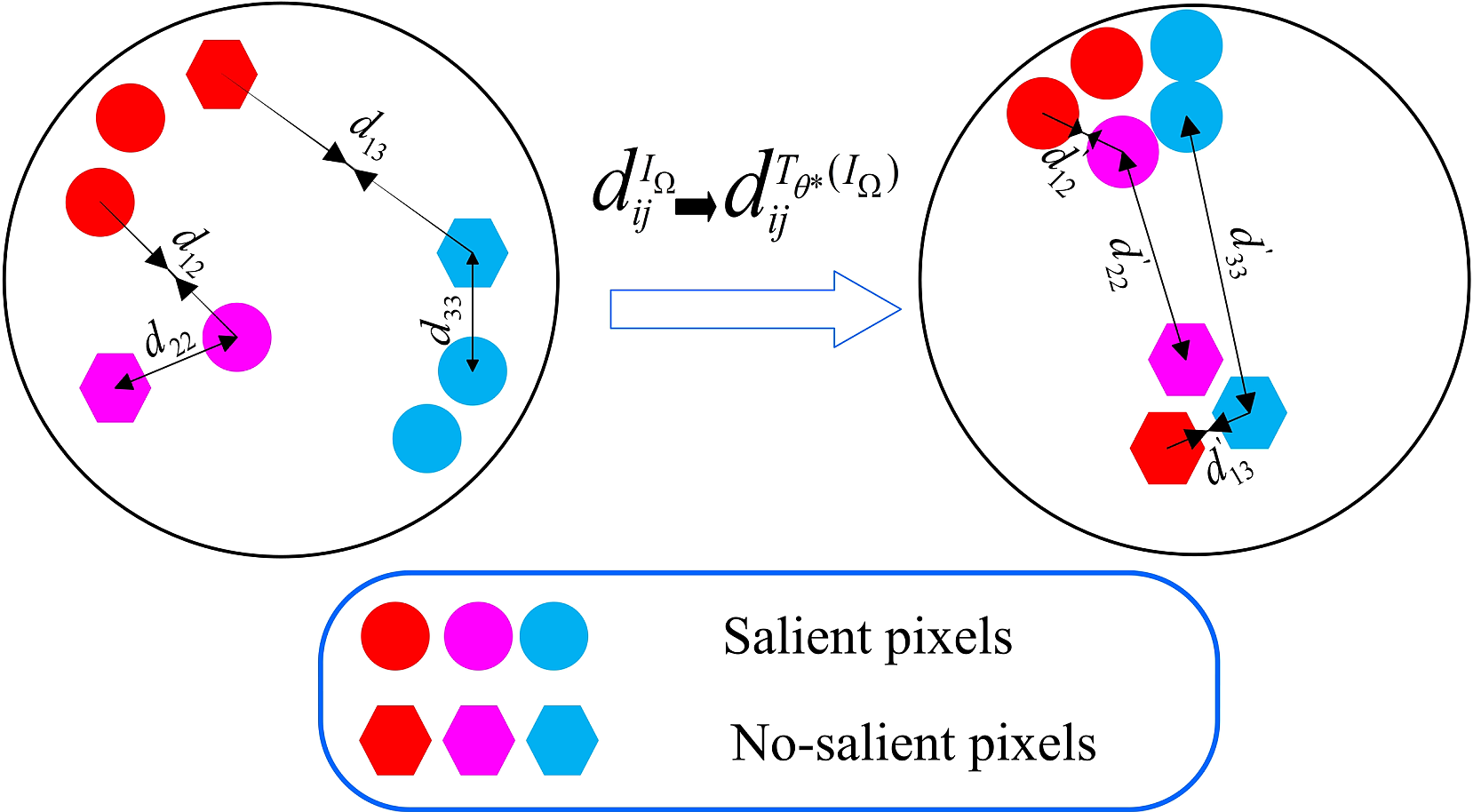}
\caption{Semantic Distance expression with the network.}
\label{fig.space}
\end{figure}

  If we train the MEnet to minimize the loss function ${L_{MEnet}}(\cdot)$, we will obtain a converged network $T_{{\theta}^*}$, where ${\theta}^*$ is the converged state of $\theta$. Given an observed input image for testing, where the pixel domain is $\Omega$, we usually describe pixel $i\in \Omega$ by its intensities $I_i$s across the channels. But it is difficult to define the semantical distance by $d_{ij}^{I_\Omega}=d(I_i,I_j)$, e.g., by Euclidean distance $d_{ij}=\|I_i-I_j\|_2$. However, through transformation of $T_{{\theta}^*}$, we will obtain the corresponding feature vectors $\{f_i\}_{i\in\Omega}$ to represent the input. Then the distance can be expressed by $d_{ij}'=d_{ij}^{T_{{\theta}^*}(I_\Omega)}=\|f_i-f_j\|_2$, and finally the saliency map $S$ for saliency segmentation is obtained by:
   \begin{equation}\label{eq.seg}
   \begin{aligned}
    S_i&=\|f_i-E_{f_j\thicksim P_B(\cdot)}f_j\|_2\\
   &={\|f_i-\sum_{j\in\Omega_B}P_B(f_j)f_j\|_2}.
   \end{aligned}
    \end{equation}
where $P_B(\cdot)$ is the probability distribution function of the feature vector $f_j \in \Omega_B$, and $\Omega=\Omega_B\cup \Omega_S$, where $\Omega_B$ and $\Omega_S$ denote the background region and salient region only computed from the component of $L_{CE}$ in the loss function (\ref{eq.7}) within the whole converged network $T_{\theta^*}$, respectively. Note that, $\Omega_B$ and $\Omega_S$  are not the accurate segmentation and they are to be further investigated in the experiment part.
To conclude, by network transformation we succeed to express $d_{ij}^{I_\Omega}$ with $d_{ij}^{T_{\theta^*}(I_\Omega)}$. As illustrated in Figure \ref{fig.space}, we anticipate that through a space transformation, the intra-class distance will be smaller than the inter-class distance.

\subsection{Noise Robustness Analysis} 
In the sense of vectors, denote the input image for the network as $x\in \mathbb{R}^n$ and output  as $y = f(x) \in \mathbb{R}^m.$ Then by differentiation, we obtain
\begin{equation}\label{eq.networkdiff}
dy_j = \frac{\partial y_j}{\partial x_i}dx_i = \frac{\partial y_j}{\partial O_{L+1}}\displaystyle \mathop{\Pi}_{l=1}^L\frac{\partial O_{l+1}}{\partial O_l}\frac{\partial O_l}{\partial x_i}dx_i,
\end{equation}
where $O_l$ denotes the output of the $l^{th}$ layer, $\frac{\partial y_j}{\partial O_{L+1}}\displaystyle \mathop{\Pi}_{l=1}^L\frac{\partial O_{l+1}}{\partial O_l}$ is the partial differentiation within back propagation. That is, we differentiate each layer's output w.r.t its corresponding input.

Firstly in the situation of 1-d output ($m=1$), by multi-dimensional Mean Value Theorem, we have
\begin{equation}\label{eq.outputdiffence}
y-\hat{y} = \nabla f(\xi) \cdot (x - \hat{x})
\end{equation}
for some vector of $\xi\in \mathbb{R}^n$ belongs to the region bounded by corresponding components of $\hat{x}$ and $x$, and the operation above is inner product. Then by Cauchy's Inequality, 
\begin{equation}|y - \hat{y}|\leq \|\nabla f(\xi)\|\cdot \|x - \hat{x}\|.
\end{equation}
Similarly, for m-d output, we have
\begin{equation}
||y - \hat{y}||\leq \|\nabla f(\xi)\|\cdot \|x - \hat{x}\|.
\end{equation}

Let $x$ and $\hat{x}$ denote the real data and its \textcolor{black}{distortion}, respectively. An assume that the \textcolor{black}{distortion} be small, $\ie \; x$ would be close enough to $\hat{x},$ then the inequality still holds for $\hat{x}$. Denote $e_{input} = x - \hat{x}$ to be the error of the input and $e_{output} = y - \hat{y}$ be the error of output, then 
\begin{equation}
    \|e_{output}\|\leq\|\nabla f(x)\|\cdot\|e_{input}\|.
\end{equation}

Thus, we can measure the robustness of $f(\cdot)$, denoted as $M$ by evaluating the norm of the {\bf Jacobian matrix} at each input data point, i.e. $M=||\nabla f(x)||$.

\subsubsection{Estimation of robustness}Suppose an arbitrary dataset $D$, then we can estimate the robustness $M$ over it, by

\begin{equation}
M=\underset{x\in D}{\mathbb{E}}\|\nabla f(x)\|
\end{equation}

where $D$ could be chosen as training, validation or test set.

\subsubsection{Approximation of Jacobian} In the case of unknown gradients, we can perform numerical differentiation to approximate the Jacobian. Since the input $f$ is multi-dimensional and without loss of generality, we assume the output be one-dimensional, which is inefficient to calculate the numerical differential directly, we can estimate it in the Monte Carlo manner w.r.t the error direction, i.e.,
\begin{equation}
\|\nabla f(x)\|_p^p = \underset{\vec{n}\in U}{\mathbb{E}} \lim_{t\xrightarrow{}0^+} \frac{|f(x+t\vec{n})-f(x)|^p}{t^p}
\end{equation}
where $U$ stands for the set of directions in which the error takes, which is assumed to be evenly distributed. Furthermore, it can adopt any distortion type of interest, such as JPEG compression, no matter it is a random error or not.

\subsubsection{Theoretical upper bound of $M$}For estimate-free analysis, we can still measure the robustness by calculating a theoretical {\bf upper bound} of the {\bf Jacobian matrix} element-wise, which is denoted as $G$. 
We then obtain the Lipschitz constant, by $M=\|G\|,$ such that $\forall \|\hat{x}-x\|$ sufficiently small, 
\begin{equation}
\| f(x)- f(\hat{x})\|\leq M\cdot \|x - \hat{x}\|
\end{equation}
In practice, \textcolor{black}{since the convolution and fully connected layer perform linear operation, the derivative should be constant with respect to its input, once the model parameters are fixed. For nonlinear operation appeared within the proposed MEnet, such as ReLU, pooling and softmax, the upper bound of derivative for arbitrary input can also be found trivially as 1, then backpropagating those upper bounds gives the total upper bound of {\bf Jacobian matrix}. Table \ref{my_gradient}} shows the direct comparison of Jacobian with other methods on selected datasets. 

\section{Experiments}
\label{experiments}

We test on several public saliency datasets and distorted images compare with state-of-the-art saliency detection methods. We use the Caffe software package to train our model\cite{jia2014caffe}. We then use Pytorch\cite{paszke2017pytorch} and Tensorflow\cite{abadi2016tensorflow} to compare MEnet with other models.

\subsection{Datasets}

The datasets we consider are: MSRA10K \cite{5}, DUT-OMRON (DUT-O) \cite{yang2013saliency}, HKU-IS \cite{li2015visual}, 
ECSSD \cite{yan2013hierarchical}, MSRA1000 (MSRA1K) \cite{liu2011learning} and SOD \cite{martin2001database}. MSRA10K contains 10000 images, it is the largest dataset and covers a large variety of contents. HKU-IS contains 4447 images, most images containing two salient objects or multiple objects. ECSSD dataset contains 1000 images. DUT-OMRON contains 5168 images, which was originally designed for image segmentation. This datasets is very challenging since most of the images contain complex scenes; existing saliency detection models have yet to achieve high accuracy on this dataset. MSRA1K including 1000 images, all belongs to the MSRA10K. 
SOD contains 300 images.

\subsection{Training}
We use stochastic gradient descent (SGD) for optimization, and the MSRA10K and HKU-IS are selected for training. For MSRA10K, 8500 images for training, 500 images for validation and the MSRA1K for testing; HKU-IS was divided into approximately 80/5/15 training-validation-testing splits. To prevent overfitting, all of our models use cropping and flipping images randomly as data augmentation. We utilize batch normalization \cite{ioffe2015batch} to steep up the convergence of MEnet.

Most experiments are performed on a PC with Intel(R) Xeon(R) CPU I7-6900k, 96GB RAM and GTX TITAN X Pascal. Some later experiments are performed on Google Colab.
We use a 4 convolutional layer block in the upsample and downsample operations. Therefore the depth of our MEnet is 52 layers. The parameter sizes are shown in Figure \ref{fig.network} and Figure \ref{fig.down-up}. \textcolor{black}{We set the learning rate to 0.1 with weight decay of $10^{-8}$, a momentum of 0.9 and a mini-batch size of 5.} We train for 110,000 iterations.
Since salient pixels and non-salient pixels are very imbalanced, network convergence to a good local optimum is challenging. Inspired by object detection methods such as SSD \cite{liu2016ssd}, we adopt hard negative mining to address this problem. This sampling scheme ensures salient and non-salient sample ratio equal to 1, eliminating label bias.\footnote {Codes and more related details are given in: \url{https://github.com/SherylHYX/Ro-SOS-Metric-Expression-Network-MEnet-for-Robust-Salient-Object-Segmentation.}}

\begin{table*}[]
\centering
\begin{tabular}{|c|c|c|c|c|c|c|c|c|c|c|c|}
\hline
Data                     & Index                  & MC    & ELD   & DHSNet & DS    & DCL      & UCF   & Amulet & NLDF     & SRM      & MEnet \\ \hline
\multirow{2}{*}{DUT-O}   & $F{_\beta}\uparrow$    & 0.622 & 0.618 & --     & 0.646 & 0.660        & 0.645 & 0.654  & 0.691    & \cwor{0.718}    & \cbet{0.732} \\ \cline{2-12}
                         & $MAE\downarrow$        & 0.094 & 0.092 & --     & 0.084 & 0.095        & 0.132 & 0.098  & 0.080    & \cbet{0.071}   & \cwor{0.074} \\ \hline
\multirow{2}{*}{HKU-IS}  & $F{_\beta}\uparrow$    & 0.733 & 0.779 & 0.859  & 0.790 & 0.844     & 0.820 & 0.841  & 0.873    & \cwor{0.877}  & \cbet{0.879} \\ \cline{2-12}
                         & $MAE\downarrow$        & 0.099 & 0.072 & 0.053  & 0.079 & 0.063    & 0.072 & 0.052  & 0.048    & \cwor{0.046}  & \cbet{0.044} \\ \hline
\multirow{2}{*}{ECSSD}   & $F{_\beta}\uparrow$    & 0.779 & 0.810 & 0.877  & 0.834 & 0.857    & 0.854 & 0.873  & \cwor{0.880}    & \cbet{0.892}   & \cwor{0.880} \\ \cline{2-12}
                         & $MAE\downarrow$        &0.106   & 0.080 & \cwor{0.060} & 0.079 & 0.078     & 0.078 & \cwor{0.060}  & 0.063 & \cbet{0.056}  & \cwor{0.060} \\ \hline
\multirow{2}{*}{MSRA1K}  & $F{_\beta}\uparrow$    & 0.885 & 0.882 & --    & 0.858 & \cwor{0.922}     & -- & --  & --         & 0.894  & \cbet{0.928} \\ \cline{2-12}
                         & $MAE\downarrow$        & 0.044 & 0.037 & --    & 0.059 & \cwor{0.035}     & -- & --  & --       & 0.045  & \cbet{0.028} \\ \hline
\multirow{2}{*}{SOD}     & $F{_\beta}\uparrow$    & 0.497 & 0.540 & 0.595 & 0.552 & 0.573  & 0.557 & 0.550  & 0.591                    & \cbet{0.617}  & \cwor{0.594} \\ \cline{2-12}
                         & $MAE\downarrow$        & 0.160 & 0.150 & 0.124 & 0.141 & 0.147  & 0.186 & 0.160  & \cwor{0.130} & \cbet{0.120}  & 0.139 \\ \hline
\end{tabular}
\caption{Comparison of quantitative results including F-measure (lager is better) and MAE (smaller is better). The top two results are indicated by \protect\cbet~ and \protect\cwor, respectively.
DHSNet is trained on MSRA-B and DUT-O, and UCF, Amulet and NLDF  are all trained on MSRA-B dataset which contains MSRA1K, therefore, we are not compared our model with this four models on this dataset. }
\label{Table_1}
\end{table*}

\begin{figure*}[!htb]
     \includegraphics[width = .95\textwidth]{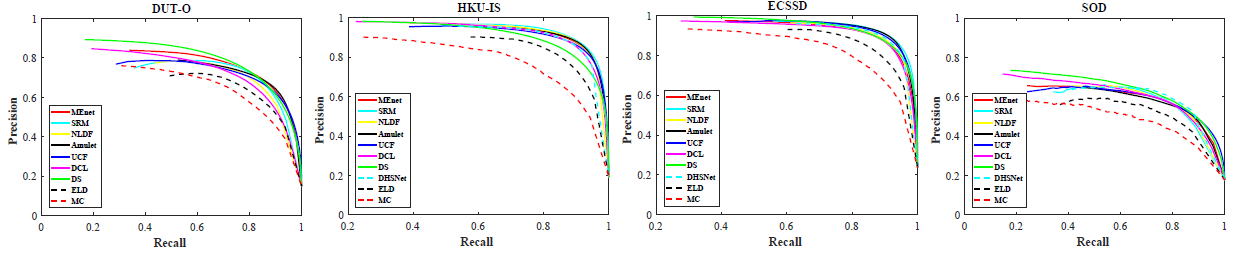}
     \caption{Comparison of precision-recall curves of other CNN-based methods on four datasets. }
 \label{fig.norm}\centering
\end{figure*}

\begin {figure*}[!t]
\centering
\includegraphics[width=0.98\textwidth]{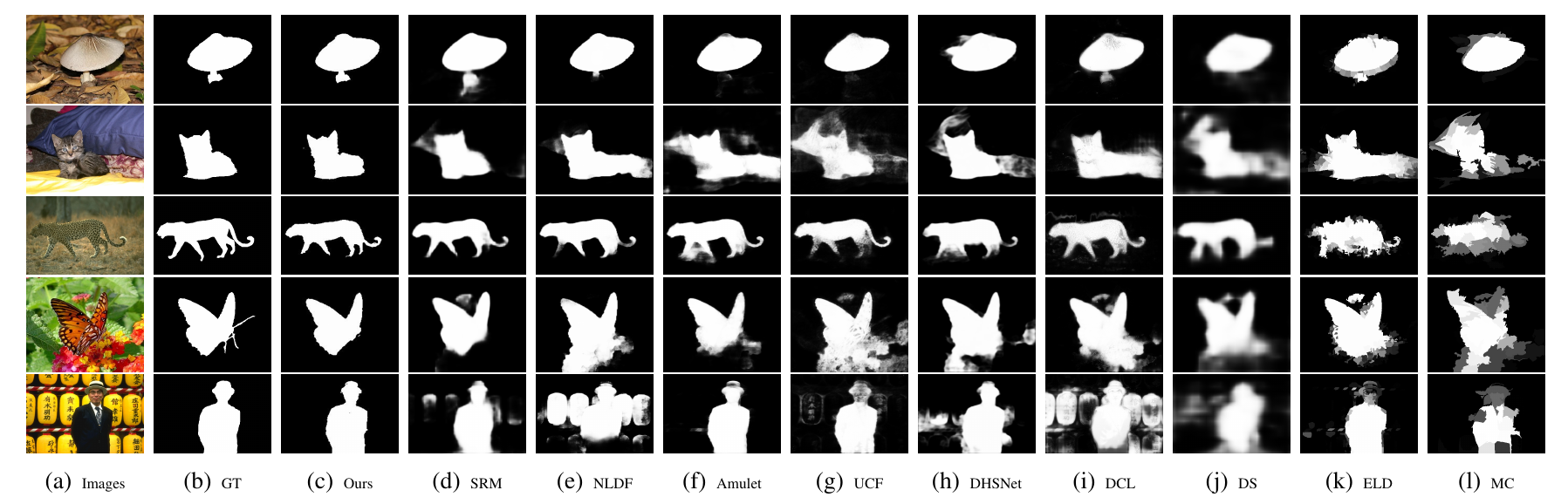}
\caption{Visual comparisons with nine methods. MEnet can obtain detailed and accurate saliency maps.}
\label{fig.comparison}
\end {figure*}

\begin{figure*}[!htb]\centering
     \includegraphics[width = .95\textwidth]{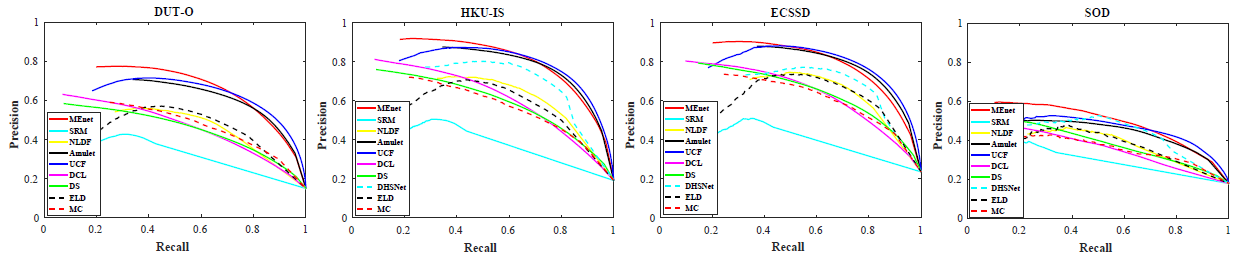}
     \caption{Comparison of precision-recall curves of other CNN-based methods on four datasets  corrupted by AWGN(with random strengths) . }
    \label{fig.pic}
\end{figure*}

\begin{table*}[!t]
\centering
\begin{tabular}{|c|c|c|c|c|c|c|c|c|c|c|c|}
\hline
Data                    & Distortion         & MC    & ELD   & DHSNet & DS    & DCL      & UCF   & Amulet & NLDF                 & SRM   & MEnet \\ \hline
\multirow{2}{*}{DUT-O}  & AWGN               & 0.487 & 0.534 & --     & 0.452 & 0.472      & \cwor{0.582} & 0.574  & 0.517 & 0.533 & \cbet{0.673} \\ \cline{2-12}
                        & Compression        & 0.512 & 0.558 & --     & 0.508 & 0.543       & 0.564 & 0.538  & \cwor{0.574} & 0.565 & \cbet{0.699} \\ \hline
\multirow{2}{*}{HKU-IS} & AWGN               & 0.554 & 0.622 & 0.723  & 0.567 & 0.589     & 0.720 & \cwor{0.726}  & 0.637 & 0.572 & \cbet{0.751} \\ \cline{2-12}
                        & Compression        & 0.602 & 0.687 & \cwor{0.736}   & 0.614 & 0.679  & 0.705 & 0.688  & 0.700 & 0.663 & \cbet{0.811} \\ \hline
\multirow{2}{*}{ECSSD}  & AWGN               & 0.603 & 0.671 & 0.726  & 0.611 & 0.618     & 0.730 & \cwor{0.738}  & 0.669 & 0.583 & \cbet{0.740} \\ \cline{2-12}
                        & Compression        & 0.650 & 0.730 & \cwor{0.753}   & 0.650 & 0.658  & 0.724 & 0.711  & 0.694 & 0.665 & \cbet{0.810} \\ \hline
\multirow{2}{*}{MSRA1K} & AWGN               & 0.750 & 0.806 & --    & 0.711 & 0.760      & --    & --     & --    & \cwor{0.785} & \cbet{0.899} \\ \cline{2-12}
                        & Compression        & 0.788 & 0.841 & --    & 0.772 & \cwor{0.833}   & --    & --     & --    & 0.820 & \cbet{0.914} \\ \hline
\multirow{2}{*}{SOD}    & AWGN               & 0.403 & 0.443 & \cwor{0.492} & 0.373 & 0.388  & 0.439 & 0.453  & 0.430 & 0.443 & \cbet{0.504} \\ \cline{2-12}
                        & Compression        & 0.410 & 0.458 & \cwor{0.475} & 0.421 & 0.406  & 0.451 & 0.431  & 0.455 & 0.433 & \cbet{0.536} \\ \hline
\end{tabular}
\caption{Quantitative comparison with recent deep methods based on deep learning methods in difference distorted scenes via F-measure (lager is better). The top two results are indicated by \protect\cbet~ and \protect\cwor, respectively.}
\label{my-label}
\end{table*}

\begin{table*}[!t]
\centering
\begin{tabular}{|c|c|c|c|c|c|c|c|c|c|c|c|}
\hline
\multirow{2}{*}{Data}  & \multicolumn{2}{|c|}{max(abs(g))}               & \multicolumn{2}{|c|}{min(abs(g))} & \multicolumn{2}{|c|}{median(abs(g))} & \multicolumn{2}{|c|}{mean(abs(g))}      & \multicolumn{2}{|c|}{var(abs(g))}  \\ \cline{2-11}
                        & MEnet        & NLDF & MEnet        & NLDF     & MEnet        & NLDF       & MEnet        & NLDF  & MEnet        & NLDF  \\ \hline
                        
DUT	&1.36E-08	&1.91E-07	&4.11E-16	&8.66E-15	&1.21E-10	&2.37E-09	&2.62E-10	&4.88E-09	&5.01E-19	&6.43E-17 \\ \hline
ECSSD	&1.04E-08	&1.84E-07	&3.34E-16	&7.24E-15	&9.70E-11	&1.96E-09	&2.05E-10	&4.36E-09	&2.43E-19	&5.68E-17 \\ \hline
HKU-IS	&7.80E-09	&1.89E-07	&2.80E-16	&7.18E-15	&7.39E-11	&2.05E-09	&1.57E-10	&4.52E-09	&1.59E-19	&5.85E-17 \\ \hline
MSRA1000	&1.12E-08	&2.17E-07	&2.77E-16	&9.02E-15	&7.76E-11	&2.35E-09	&1.90E-10	&5.21E-09	&3.24E-19	&8.27E-17 \\ \hline
SOD	&1.22E-08	&1.75E-07	&4.79E-16	&6.87E-15	&1.24E-10	&1.97E-09	&2.56E-10	&4.21E-09	&3.32E-19	&5.11E-17 \\ \hline

\end{tabular}
\caption{\textcolor{black}{Direct Jacobian comparison with state-of-the-art deep learning method. Specifically, g is the gradient of the sum of output with respect to the input for dimension reduction and the statistical criterion (the max, min, median, mean, var) is performed with each image, which is then averaged to produce final result for the corresponding dataset. MEnet shows superior performance on all criteria of all datasets considered.}}
\label{my_gradient}
\end{table*}

\begin{figure}[t!]
\centering
    \includegraphics[width=0.98\columnwidth]{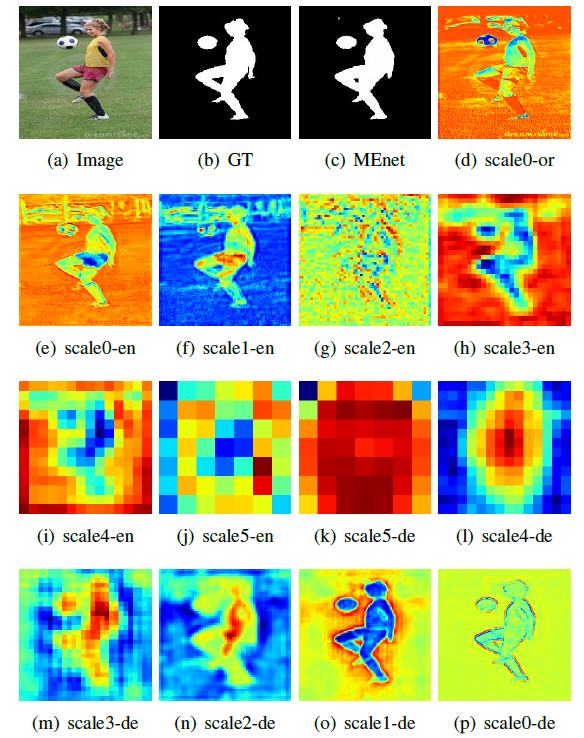}
     \caption{Feature maps visualization, where (d) is the feature of original data, (e)-(j) and (k)-(p)denote the learned feature maps of encoder and decoder, respectively. }
    \label{fig.feature}
  \end{figure}


   \begin{figure*}[t]
  \centering
\includegraphics[width=0.95\textwidth]{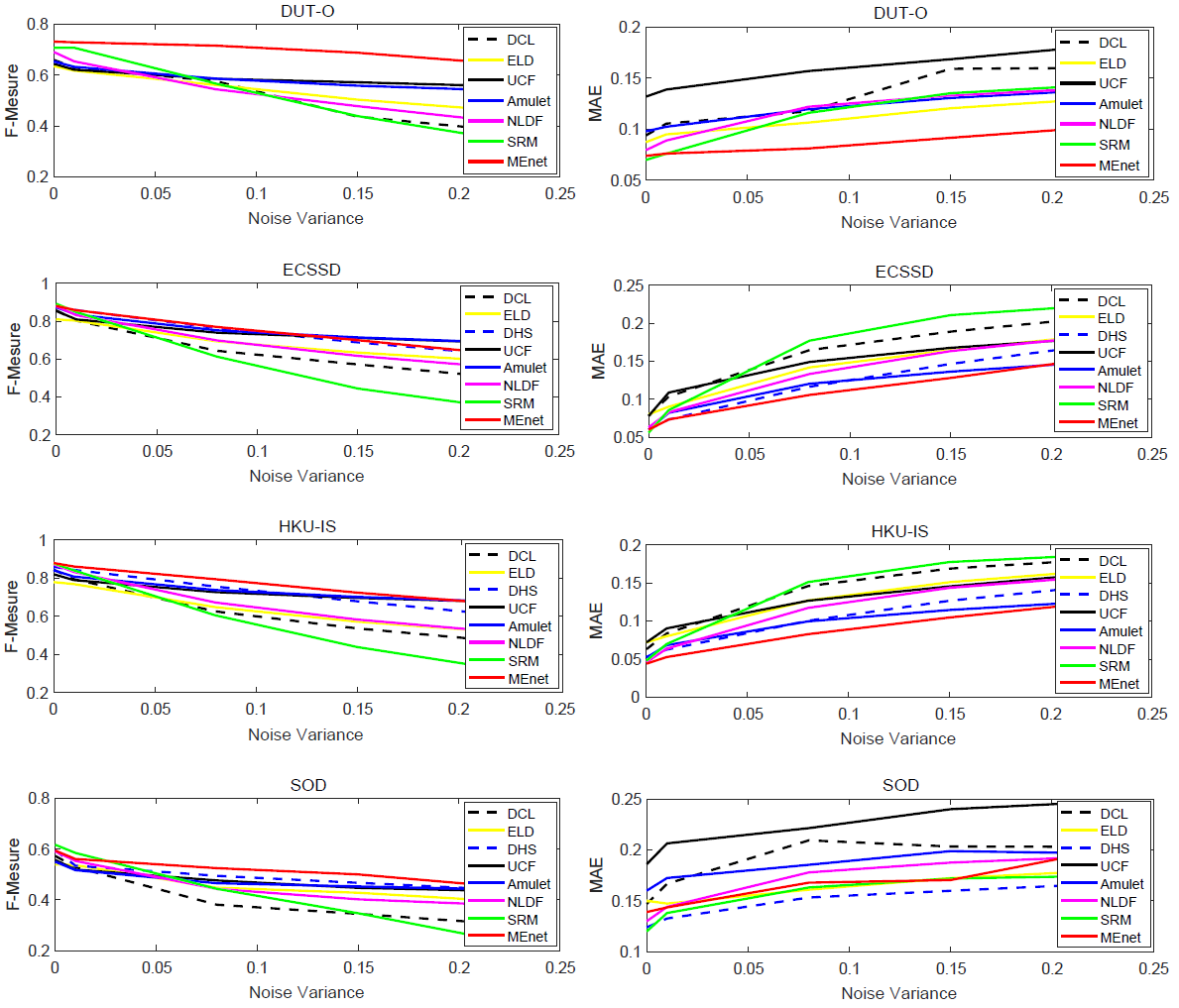}
\caption{The curves of difference methods on public datasets under various noise variances.} 
\label{fig.loss_curve}
\end{figure*}

\begin{table}[t]
\small
\centering
 \begin{tabular}{|c|c|c|c|c|c|c|}
 \hline
 Data                   &	Indexes	          	&CE-plain      & CE-only                    & MEnet \cr
 \hline
 \multirow{2}{*}{DUT-O} &$F{_\beta}\uparrow$    &0.631   & 0.678                   & $\bf{0.732}$  \cr\cline{2-5}
                        & $MAE\downarrow$       &0.098   &0.084                   &$\bm{0.074}$ \cr\cline{1-5}
 \multirow{2}{*}{HKU-IS}&$F{_\beta}\uparrow$    &0.803   & 0.872                  &$\bf{0.879}$  \cr\cline{2-5}
                        & $MAE\downarrow$       &0.064   &0.056                   &$\bm{0.044}$ \cr\cline{1-5}
 \multirow{2}{*}{ECSSD} &$F{_\beta}\uparrow$    &0.794   & 0.855                  &$\bf{0.880}$  \cr\cline{2-5}
                        & $MAE\downarrow$       &0.093   &0.072                   &$\bm{0.060}$ \cr\cline{1-5}
 \multirow{2}{*}{MSRA1K}&$F{_\beta}\uparrow$    &0.884   & 0.915                 &$\bf{0.928}$  \cr\cline{2-5}
                        & $MAE\downarrow$       & 0.037  &0.034                  & $\bm{0.028}$\cr\cline{1-5}
 \multirow{2}{*}{SOD}  &$F{_\beta}\uparrow$     & 0.525   & 0.555                 &$\bf{0.594}$  \cr\cline{2-5}
                       & $MAE\downarrow$        &  0.156  &0.159                 & $\bm{0.139}$\cr\cline{1-5}
\end{tabular}
\caption{The performance of different strategies.}
\label{Table 3}
\end{table}
\subsection{Upsampling Operation}
\textcolor{black}{In our approach, we use two upsampling operation methods. As shown in Figure \ref{fig.feature}, we use deconvolution \cite{long2015fully} as the upsampling method in Decoder operation. 
Since  feature maps are generated at different scales, we use a simple way to upscale output feature maps with different sizes for concatenation. The response of each upscale operation is defined as:
\begin{equation}\label{concatenation}
    f^{l+1}(nx-n+1 :nx,ny-n+1:ny)=f^{l}(x,y)
    \end{equation}
where $x$ and $y$ denote the input and output respectively; $l$ is the convolutional level, and $n\times n$ is the size of input.}

\subsection{Quantitative evaluation}

\textcolor{black}{For the valuation of the result, we use three standard criteria evaluation method: F-measure rate for adaptive threshold, mean absolute error (MAE) \cite{borji2015salient,23}, and precision-recall curve (PR curve) \cite{PRcurve}. PR curve is widely utilized, and is obtained by comparing result with the ground truth by the binary masks generated with the threshold sliding from 0 to 255. For the F-measure, the adaptive threshold $T_{adp}$ \cite{adaptiveT} is defined as twice the mean saliency of the image as shown in Equation \ref{eq.t}.}

\textcolor{black}{\begin{equation}\label{eq.t}
T_{adp}=\frac{2}{W\times H}\sum_{x=1}^{W}\sum_{y=1}^{H}S(x,y)
\end{equation}
where $W$ and $H$ denote the width and the height of the final saliency map S respectively. And the F-measure is defined as
\begin{equation}\label{eq.F}
F_{\beta}=\frac{(1+\beta^{2})\cdot Precision\cdot Recall}{\beta^{2}\cdot Precision+Recall}
\end{equation}
where $\beta^{2}=0.3$. Different from PR curves, MAE evaluates the averaged degree of the dissimilarity and comparability between the saliency map and the ground truth at every pixel.
MAE is defined as the average pixelwise absolute difference between the binary ground truth and the saliency map.
\begin{equation}\label{eq.MAE}
MAE = \frac{1}{W\times H}\sum_{x=1}^{W}\sum_{y=1}^{H}|S(x,y)- G(x,y)|
\end{equation}}
\subsection{Performance Comparison}
We compare MEnet with 9 state-of-the-art models for saliency detection: MC \cite{MC}, ELD \cite{ELD}, DCL \cite{li2016deep}, DHSNet \cite{DHSNet}, DS \cite{DS}, UCF \cite{UCF}, Amulet \cite{Amulet}, SRM \cite{SRM}, NLDF \cite{luo2017non} and 2 traditional metric learning methods: AML \cite{li2015adaptive} and Lu's method \cite{you2016salient}.

\subsubsection{Visual Comparison}

\textcolor{black}{A visual comparison is shown in Figure \ref{fig.comparison} along with other state-of-the-art methods. To illustrate the efficiency of MEnet, we select a variety of difficult circumstances from different datasets. MEnet performs better in these challenging circumstances e.g., when the salient region is similar to background and the images with complex scenes.}

\subsubsection{F-measure and MAE}

\textcolor{black}{We also compare our approach with the state-of-the-art salient segmentation methods in terms of F-measure scores and MAE are shown in Table \ref{Table_1}.} \textcolor{black}{It is noted that the better models (e.g., DHSNet, NLDF, Amulet, SRM, UCF and etc.) need pre-trained model, and conditional random field (CRF) method \cite{krahenbuhl2011efficient} is used as post-processing in DCL. Although MEnet is trained from scratch, it is still comparable with state-of-the-art models (average F-measure with  adaptive threshold and MAE), particularly on some challenging datasets DUT-O and HKU-IS.}
Our model costs $0.086s$ to generate each saliency map with GPU.



\subsubsection{PR curve}

\textcolor{black}{We compare our approach with existing methods by the PR curve approach which is widely used to evaluate the performance of salient segmentation models. As shown in Figure \ref{fig.norm}, since MSRA5K which contains MSRA1K is treated as the training dataset, we only depict the PR curves produced by our approach and other previous methods on four datasets. From Figure \ref{fig.norm}, it's clear that MEnet is comparable with state-of-the-art models without any pre/post processing.}

\begin{figure*}[!t]
     \centering
          \includegraphics[width=0.98\textwidth]{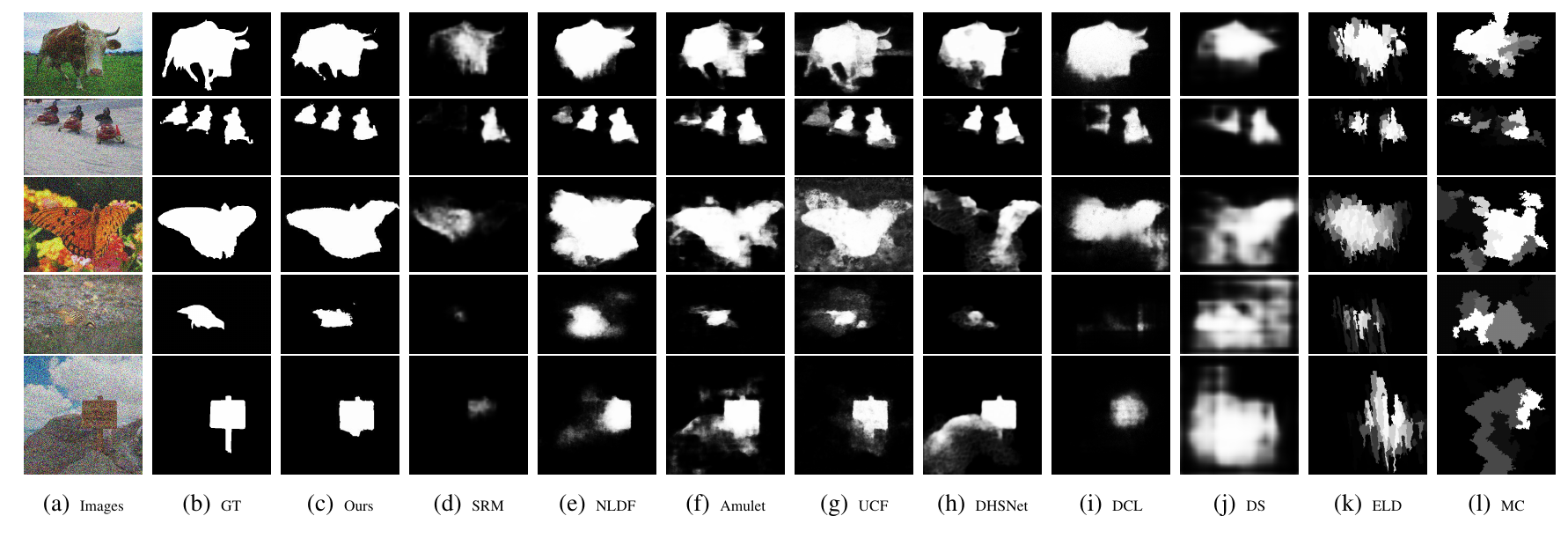}
     \caption{Visual comparisons with nine methods on public datasets with AWGN. MEnet can obtain detailed and accurate saliency maps. }
    \label{fig.AWGN}
  \end{figure*}

\begin{figure*}[!t]
     \centering
          \includegraphics[width=0.98\textwidth]{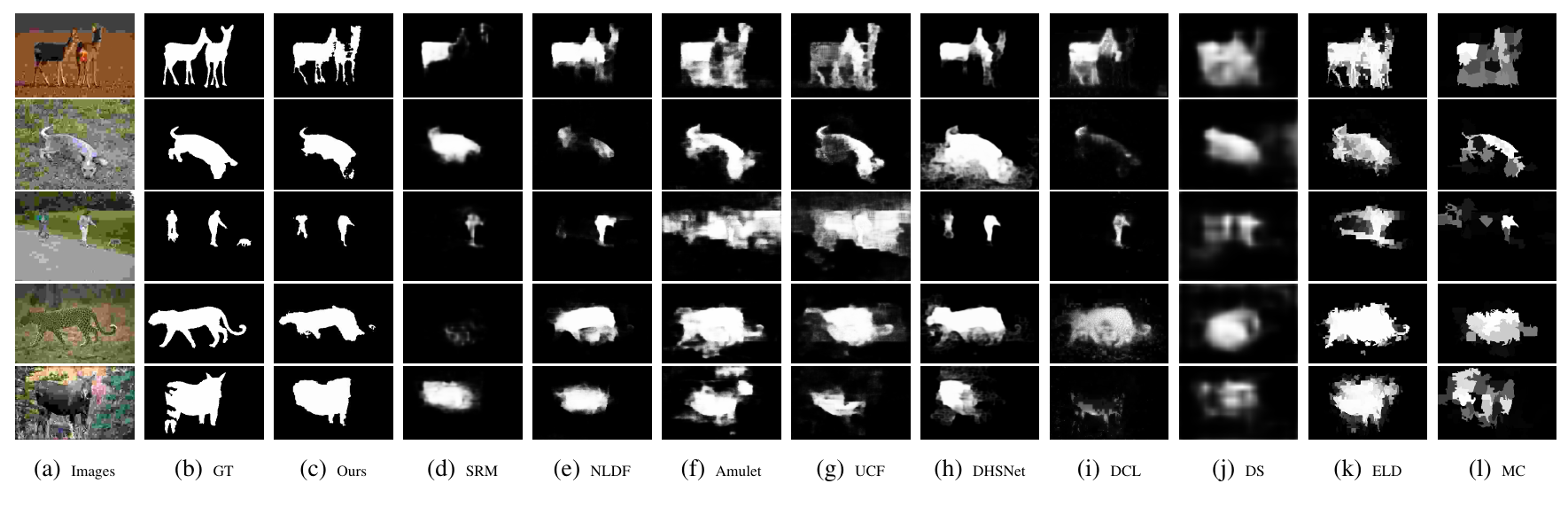}
     \caption{Visual comparisons with nine methods on public datasets with JPEG Compression. MEnet can obtain detailed and accurate saliency maps. }
    \label{fig.Compression}  
  \end{figure*}

\textcolor{black}{\subsection{Robustness evaluation}}
\textcolor{black}{Note that MEnet is not trained on distorted images, which is the same as previous works.} To show the robustness of MEnet in the distorted setting, we work with public datasets corrupted by Additive White Gaussian Noise (AWGN) and JPEG compression (with random strengths). 
\textcolor{black}{We compare F-measure scores in Table \ref{my-label}. We can see that MEnet clearly outperforms other methods. Additionally, we show PR curves of our approach in Figure \ref{fig.pic}. Since the saliency maps generated by metric loss prediction tend to be binary, it is harmful to draw PR curves which need continuous salient values. Therefore, we select saliency maps generated by CE prediction to draw PR curves. Through Figure \ref{fig.pic}, we observe that the performance of the proposed method is a little better than others on distorted datasets.  As shown in Figure \ref{fig.loss_curve}, with growing noise variance, the performance of other methods degrades rapidly, while MEnet still achieves robust performance.} The reason for the robustness of MEnet owes to the fact that multi-scale features and metric loss are integrated into this structure, where abundant features from either low or high levels are fully utilized and metric loss idea correlates every pixel to the remaining pixels for optimization. For example, a similar metric loss idea is shown to be robust in human re-identification \cite{yi2014deep}, because it is insensitive to light, deformation and angle that can be regarded as ``noise''. In practice, images are easy to be impacted by noise and compression. Therefore, our proposed work is beneficial for constructing a robust model.
\textcolor{black}{\subsubsection{Evaluation on distorted images}To better explain the efficiency of MEnet on distorted images, we also compare our approach with the state-of-the-art methods in qualitative evaluation. We select several images which are disturbed by AWGN or JPEG compression with different parameters, as Figure \ref{fig.AWGN} and Figure \ref{fig.Compression} demonstrate. From Figure \ref{fig.AWGN}, we can observe that MEnet also can obtain the accurate saliency maps with small variance as shown in the top three rows. From the bottom two rows, we can see that MEnet also outperforms other methods. Figure \ref{fig.Compression} illustrates that for the images disturbed by JPEG compression, MEnet can detect large parts of the accurate salient objects, which is better than other previous methods..   
\subsubsection{Jacobian of test datasets}To further illustrate the robustness of MEnet, we compare the Jacobians on several datasets that are not trained on. Since the Jacobian is given with respect to each input and output pixel, criteria of dimensional reduction are applied. As can be observed in Table \ref{my_gradient}, MEnet outperforms other methods by a magnitude in all criteria. For this comparison, MEnet is performed on PyTorch and NLDF is on TensorFlow.
}

\begin{table}[t]
\centering
 \begin{tabular}{|c|c|c|c|c|c|}
\hline
	Data &	Indexs	&	AML
	&	Lu's & MEnet \cr
\hline
\multirow{2}{*}{ECSSD} &$F{_\beta}\uparrow$ &0.667 & 0.715 &0.880  \cr\cline{2-5}
& $MAE\downarrow$ & 0.165 &0.136 &0.060 \cr\cline{1-5}
\multirow{2}{*}{MSRA1K} &$F{_\beta}\uparrow$ &0.794 & 0.806 &0.928  \cr\cline{2-5}
& $MAE\downarrow$& 0.089 &0.080 & 0.028\cr\cline{1-5}
\end{tabular}
 \caption{Comparison with two traditional methods based on metric learning with F-measure and MAE scores.}
\label{Table 2}
\end{table}



%

 \subsection{Advantages of MEnet}

To intuitively illustrate the advantages of MEnet, we select several feature maps visualization for analysis. As the layers of each scale go deeper, the receptive field of each neuron becomes larger.   As shown in Figure \ref{fig.feature}, we observe that each convolutional layer contains different semantic information, and going deeper allows the model to capture richer structures. Within the decoding parts, scale2-de, 3-de, 4-de are sensitive to the salient region, while scale1-de has higher response against the background region. Other layers like scale0-de can distinguish the boundary of salient objects.

To show the effectiveness of our proposed multi-scale feature extraction and loss function, we use different strategies for semantic saliency detection/segmentation as shown in Table \ref{Table 3}. 
\textcolor{black}{ The difference between CE-only and CE-plain is that CE-plain does not utilize multi-scale information which will cause performance degradation. }
We also note that the performance of MEnet is improved after introducing metric loss.

 We compare MEnet with two other traditional metric learning methods for saliency segmentation, AML \cite{li2015adaptive} and Lu \cite{you2016salient}. The results in Table \ref{Table 2} demonstrate the potential superiority of deep metric learning method over traditional metric learning for semantic saliency detection.

\begin{figure}[!t]
\centering
     \includegraphics[width = 0.49\textwidth]{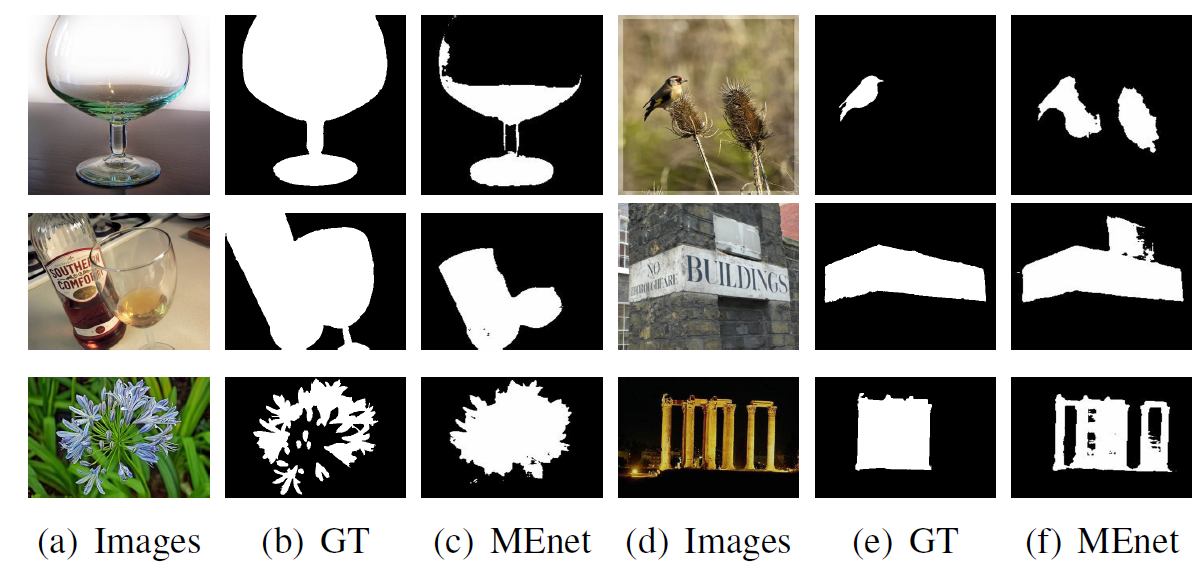}
  \caption{Failure cases selected from public datasets.}
    \label{fig.bad}
  \end{figure}

\subsection{Failure Cases}

\textcolor{black}{However, as shown in Figure \ref{fig.bad}, MEnet may also sometimes mistake information as a salient object based on the provided ground truth. Simultaneously, we observe that our approach can sometimes fail to detect some parts of transparent object, so there is still room to improve. However, in these cases, we see that the mistakes are often semantically reasonable, rather than caused by a clear flaw in the model.}
\section{Conclusion}
An end-to-end deep metric learning architecture (called MEnet) for salient object segmentation is presented in this paper. Within this architecture, multi-scale feature extraction is utilized to obtain adequate semantic information which is then combined with deep metric learning. Our network maps pixels into a ``saliency space'' for Euclidean metrics to be measured. The mapping result in the ``saliency space'' effectively distinguishes salient image elements (pixels) from background. Besides, MEnet is trained from scratch and does not require pre/post-processing. Experimental results on benchmark datasets demonstrate the outstanding performance of our model. Comparison with existing methods on distorted images and numerical evaluation of robustness show the robustness of our model to distortion.

\bibliography{refs}
\bibliographystyle{ieeetr}

\end{document}